\documentclass[sigconf]{acmart}
\AtBeginDocument{%
  }
    
\usepackage{multirow}
\usepackage{makecell}

\setcopyright{acmlicensed}
\copyrightyear{2026}
\acmYear{2026}
\acmDOI{XXXXXXX.XXXXXXX}
\acmConference[preprint]{}
\acmISBN{in progress}

\begin{document}

\title{SurveyEval: Towards Comprehensive Evaluation of LLM-Generated Academic Surveys}

\author{Jiahao Zhao}
\authornote{Both authors contributed equally to this research.}
\email{zhaojiahao2019@ia.ac.cn}
\affiliation{%
  \institution{Beijing Wenge Technology Co., Ltd. }
  \institution{Institute of Automation, Chinese Academy of Sciences}
  \city{Beijing}
  \country{China}
}

\author{Shuaixing Zhang}
\authornotemark[1]
\email{shuaixing.zhang@wenge.com}
\affiliation{%
  \institution{Beijing Wenge Technology Co., Ltd. }
  \city{Beijing}
  \country{China}
}

\author{Nan Xu}
\email{nan.xu@wenge.com}
\affiliation{%
  \institution{Beijing Wenge Technology Co., Ltd. }
  \city{Beijing}
  \country{China}
}

\author{Lei Wang}
\email{lei.wang@wenge.com}
\affiliation{%
  \institution{Beijing Wenge Technology Co., Ltd. }
  \city{Beijing}
  \country{China}
}

\renewcommand{\shortauthors}{Zhao et al.}

\begin{abstract}
LLM-based automatic survey systems are transforming how users acquire information from the web by integrating retrieval, organization, and content synthesis into end-to-end generation pipelines. While recent works focus on developing new generation pipelines, how to evaluate such complex systems remains a significant challenge. To this end, we introduce \textbf{SurveyEval}, a comprehensive benchmark that evaluates automatically generated surveys across three dimensions: overall quality, outline coherence, and reference accuracy. We extend the evaluation across 7 subjects and augment the LLM-as-a-Judge framework with human references to strengthen evaluation–human alignment. Evaluation results show that while general long-text or paper-writing systems tend to produce lower-quality surveys, specialized survey-generation systems are able to deliver substantially higher-quality results. We envision SurveyEval as a scalable testbed to understand and improve automatic survey systems across diverse subjects and evaluation criteria.
\end{abstract}





\maketitle

\section{Introduction}
The rapid advancement of large language models (LLMs) has demonstrated remarkable potential in complex text-generation tasks such as academic writing, literature reviews, and scientific reports \cite{agarwal2024litllms}. LLM-based automatic survey systems are transforming how users acquire knowledge from vast information repositories by integrating retrieval, organization, and content synthesis into streamlined end-to-end generation pipelines \cite{ali2024automatedLiteratureReview}.

Both academia and industry have introduced various specialized systems for survey generation. These automated writing systems can be broadly categorized into three types: general long-text writing systems (e.g., Kimi~\cite{Kimi_2025}, GLM~\cite{GLM_2025}) that provide broader capabilities for extended text generation; paper-writing systems (e.g., Chengpian~\cite{Chengpian_CP_2025}, Doubao Paper Mode~\cite{Doubao_2025}) that focus on structured composition of complete research papers; and survey-writing agents (e.g., SurveyGo~\cite{surveygo}, SurveyX~\cite{surveyx}, Panshi ScienceOne~\cite{ScienceOne}) specifically designed for academic survey generation. These systems not only process large volumes of literature and extract key information, but also generate well-structured survey drafts complete with sectioned organization, citation annotations, and logical coherence.

However, while recent research has primarily focused on developing new generation pipelines, how to evaluate such complex systems remains a significant challenge. Existing evaluation approaches often rely on ad-hoc human subjective scoring of individual cases, lacking reusable quantitative metrics that could support cross-system comparison, performance attribution, and systematic improvement. These issues severely constrain quality assurance and capability enhancement of survey-writing systems. Therefore, establishing a standardized evaluation benchmark for survey-writing systems is not only critical for ensuring output quality and reliability, but also provides the research community with a unified foundation for performance comparison and capability diagnosis.

\begin{figure*}[t]
\centering
\includegraphics[width=\linewidth]{surveybench_v4.pdf}
\caption{The overall framework of SurveyEval, which evaluates LLM-generated surveys across seven domains through content quality, outline coherence, and reference accuracy.}
\label{fig:frame}
% \vspace{-0.3cm}
\end{figure*}

To address this gap, we introduce SurveyEval, a comprehensive benchmark that evaluates automatically generated surveys across three dimensions: \textit{overall quality}, \textit{outline coherence}, and \textit{reference accuracy}. We extend the evaluation across seven academic disciplines and enhance the LLM-as-a-Judge framework with human-written reference surveys to increase alignment consistency between automated evaluation and human judgments. Our evaluation of representative automatic survey systems shows that general LLM-based writing models often yield suboptimal survey quality, whereas dedicated survey-generation systems demonstrate notably stronger performance with clearer strengths and more stable output. The main contributions are as follows:

\begin{itemize}

\item We propose SurveyEval, a comprehensive benchmark with three evaluation dimensions enhanced by human references to improve evaluation-human alignment.
\item We construct a cross-disciplinary test dataset spanning seven academic subjects with manually curated evaluation tasks.
\item We conduct systematic evaluation of representative automatic survey systems, revealing their capabilities, strengths, and limitations.

\end{itemize}

\section{SurveyEval Benchmark}
\subsection{Dataset Construction}
To enable fair evaluation across tasks and systems, we construct a multi-topic, multi-disciplinary test dataset:

\noindent\textbf{Computer Science (20 topics).} We curate topics covering core areas including large language model training, inference mechanisms, model compression, and alignment methods, balancing research frontiers with verifiability. This dataset has been previously used as test data in SurveyX~\cite{surveyx} system.

\noindent\textbf{Six STEM Disciplines (18 topics).} We extend coverage to mathematics, physics, biology, chemistry, aerospace, and geography, addressing the gap where existing datasets predominantly focus on computer science. These topics are derived from recent publications in \textit{Nature} over the past two years.

Each topic includes: (1) automatically generated surveys from target systems; (2) human-authored high-quality reference surveys serving as scoring anchors for the Human Benchmark; and (3) aligned reference literature collections for citation verification.

Table~\ref{tab:dataset_stats} presents detailed statistics of our constructed dataset across all disciplines.

\begin{table}[t]
\centering
\caption{Statistics of the SurveyEval dataset.}
\label{tab:dataset_stats}
\begin{tabular}{lccc}
\toprule
\textbf{Discipline} & \textbf{\# Topics} & \textbf{Avg. \# Sects} & \textbf{Avg. \# Refs} \\
\midrule
Computer Science & 20 & 34.7 & 220.2 \\
Astronomy & 3 & 19.0 & 99.3 \\
Biology & 3 & 18.0 & 122.7 \\
Chemistry & 3 & 31.7 & 48.7 \\
Geography & 3 & 18.3 & 147.3 \\
Aerospace & 3 & 30.3 & 77.7 \\
Physics & 3 & 30.7 & 128.3 \\
\midrule
Total & 38 & 26.1 & 120.6 \\
\bottomrule
\end{tabular}
\vspace{-2em}
\end{table}

\begin{table*}[t]
\centering
\caption{Main evaluation results on overall quality of LLM-generated survey (1--5 scale; higher is better).}
\vspace{-0.5em}
\label{tab:main_results}
\resizebox{\linewidth}{!}{
\begin{tabular}{llccccccccc}

\toprule
\makecell[c]{\textbf{Dataset}} & 
\makecell[l]{\textbf{System}} & 
\makecell[c]{\textbf{Coverage}} & 
\makecell[c]{\textbf{Structure}} & 
\makecell[c]{\textbf{Relevance}} & 
\makecell[c]{\textbf{Synthesis}} & 
\makecell[c]{\textbf{Critical} \\ \textbf{Analysis}} & 
\makecell[c]{\textbf{Veracity}} & 
\makecell[c]{\textbf{Originality} \\ \textbf{Proportion}} & 
\makecell[c]{\textbf{Depth of} \\ \textbf{Content}} & 
\makecell[c]{\textbf{Avg.}} \\
\midrule

\multirow{7}{*}{\makecell[c]{Computer \\ Science}} & Kimi & 3.10 & 3.45 & 4.30 & 3.10 & 3.30 & 3.30 & 2.20 & 3.30 & 3.26 \\
& GLM & 4.05 & 3.45 & 4.70 & 3.25 & 3.40 & 2.95 & 1.95 & 3.10 & 3.36 \\
& Doubao & 3.90 & 3.65 & 4.60 & 3.35 & 3.65 & 2.90 & 2.40 & 3.20 & 3.46 \\
& Chengpian & 2.45 & 2.85 & 3.45 & 2.45 & 2.30 & 2.40 & 2.00 & 2.35 & 2.53 \\
& SurveyX & 2.61 & 3.04 & 3.66 & 2.95 & 2.90 & 2.80 & 2.23 & 2.00 & 2.77 \\
& SurveyGo & 4.90 & 3.95 & 4.95 & 4.05 & 4.00 & 3.50 & 2.90 & 3.70 & 3.99 \\
& ScienceOne & 4.50 & 4.30 & 5.00 & 3.95 & 4.25 & 3.90 & 3.05 & 4.15 & 4.14 \\
\midrule
\multirow{7}{*}{\makecell[c]{Six STEM \\ Disciplines}} & Kimi & 3.94 & 4.16 & 4.66 & 3.55 & 3.22 & 3.00 & 2.83 & 3.27 & 3.58 \\
& GLM & 3.77 & 3.77 & 4.27 & 3.05 & 2.44 & 3.00 & 2.16 & 2.94 & 3.18 \\
& Doubao & 4.16 & 4.16 & 4.72 & 3.55 & 3.00 & 2.88 & 2.50 & 3.22 & 3.52 \\
& Chengpian & 2.72 & 3.05 & 3.44 & 2.55 & 2.16 & 2.55 & 2.27 & 2.33 & 2.63 \\
& SurveyX & 3.20 & 3.00 & 3.55 & 2.88 & 2.77 & 2.66 & 1.88 & 2.16 & 2.76 \\
& SurveyGo & 4.94 & 4.50 & 4.94 & 4.33 & 4.44 & 4.27 & 3.11 & 4.16 & 4.34 \\
& ScienceOne & 4.83 & 4.61 & 4.94 & 4.16 & 4.44 & 4.00 & 3.66 & 4.27 & 4.36 \\
\bottomrule
\end{tabular}
}
\vspace{-0.5em}
\end{table*}

\subsection{Evaluation Metrics}

To evaluate automatically generated surveys, we adopt two complementary approaches. For content quality, we use an enhanced LLM-as-a-Judge framework that leverages human-written reference surveys to improve alignment with expert judgment. For outline coherence, we apply a principle-based LLM assessment without reference materials. For reference quality, we rely on standard citation metrics (Recall, Precision, F1).

\paragraph{\textbf{Overall}}  We first replicate five widely used content dimensions: \textit{coverage}, \textit{structure}, \textit{relevance}, \textit{synthesis}, and \textit{critical analysis} from frameworks such as SurveyForge~\cite{yan2025surveyforge}, SurveyX~\cite{surveyx}, and AutoSurvey~\cite{2024autosurvey} to establish a baseline. Beyond these, we extend content evaluation with three complementary sub-dimensions: 

\textit{Veracity}, evaluating the clear separation of facts, opinions, and hypotheses and the precision/temperance of terminology with evidence-backed assertions; 

\textit{Originality Proportion}, assessing the proportion and quality of original contributions (e.g., novel taxonomies, critical insights, and forward-looking directions) beyond mere aggregation; 

\textit{Depth of Content}, measuring logical depth and structured synthesis that distinguish shallow collection from well-organized, reasoned analysis. 

All sub-dimensions use a unified 1–5 scale (5 best).

\paragraph{\textbf{Outline}} We evaluate the outline as a first-class objective because it determines the structural skeleton and organizational logic of a survey. The assessment targets three dimensions: 

(i) Structural Organization (Struc.): adherence to field-appropriate schemas (e.g., IMRaD, chronological development, method-based taxonomy) that support clear information localization and rapid retrieval.

(ii) Content Value (Cont.): inclusion of meaningful scholarly elements such as research gaps, conceptual frameworks, synthesized trends, limitations, and future directions, instead of simple enumeration.

(iii) Descriptiveness (Desc.): use of precise and concise section titles that correctly reflect scope and entities while avoiding vague or overly broad phrasing.

All dimensions use a unified 1–10 scale (10 best).

\paragraph{\textbf{Reference}} We evaluate citation quality using standard metrics: Citation Recall, Citation Precision, and their harmonic mean (F1). Recall measures how many ground-truth references (from the human-written survey) are correctly included in the generated survey, reflecting the breadth of literature coverage. Precision assesses whether the references cited in the generated survey indeed appear in the ground-truth reference set, indicating citation accuracy rather than arbitrary or hallucinated sources. In this way, we directly compare the generated reference list with the human reference list, offering a clear and interpretable view of how closely an automatic system’s literature grounding aligns with human research practice. This protocol allows systematic comparison of the depth and breadth of literature investigation between automatic survey systems and human-written surveys.

\subsection{LLM-as-a-Judge}

LLM-as-a-Judge has become a widely adopted paradigm for evaluating generated text, owing to its scalability, consistency, and low annotation cost. In its standard form, a judging model is prompted with rubric definitions and score ranges, and assigns quality ratings to system outputs. Despite its convenience, this rubric-only setup often exhibits leniency and score saturation, particularly for high-level generation tasks such as long-text surveys. To better accommodate the distinct characteristics of our evaluation dimensions, we employ two complementary variants of LLM-as-a-Judge: one \emph{with human-written reference} for overall content evaluation, and one \emph{with principle} for outline evaluation.

\subsubsection{LLM-as-a-Judge with Human-Written Reference}

For overall survey quality assessment, we adopt an enhanced configuration in which the judging model is provided with a human-written survey on the same topic alongside the generated text. The human-written survey acts as a high-quality anchor, enabling the model to form relative judgments rather than relying solely on abstract rubric descriptions.

This reference-guided setup mitigates several limitations of rubric-only evaluation. First, the presence of a concrete expert example offers strong contextual anchoring, reducing the tendency of the judging model to overestimate vague, generic, or shallow content. Second, comparing the generated survey against an analytically rich human counterpart increases the model’s sensitivity to higher-level qualities such as argumentative rigor, evidential grounding, and conceptual contribution—areas where rubric-only scoring often becomes overly permissive. Third, anchoring the judgment to a gold-standard sample alleviates the score saturation frequently observed when different systems receive uniformly high marks despite substantial quality differences. As a result, this method yields clearer separation among systems and provides a more reliable signal for diagnosing strengths and weaknesses in automated survey generation.

\subsubsection{LLM-as-a-Judge with Principle}

For outline evaluation, we adopt a principle-based setup without human references. Because outlines are much shorter than full surveys and encode only structural intent, comparing them directly with a human-written exemplar would be overly rigid and provide limited discriminative value. Instead, the judging model assesses each outline against explicit criteria of structural organization, logical progression, and pragmatic clarity. This design better reflects the nature of outline quality, which depends primarily on adherence to well-established scholarly conventions. The principle-based method therefore provides a more stable, fair, and interpretable evaluation tailored to the structural characteristics of outlines.

\section{Experiments}

\begin{table}[t]
\centering
\caption{Outline evaluation results. “Struc.”, “Cont.”, “Desc.” stand for structure, content value, and descriptiveness.}

\label{tab:outline_results}
\begin{tabular}{llrrrr}
\toprule
\multicolumn{1}{c}{\textbf{Dataset}} & \multicolumn{1}{l}{\textbf{System}} & \multicolumn{1}{c}{\shortstack{\textbf{Struc.}}} & \multicolumn{1}{c}{\shortstack{\textbf{Cont.}}} & \multicolumn{1}{c}{\shortstack{\textbf{Desc.}}} & \multicolumn{1}{c}{\shortstack{\textbf{Total}}} \\
\midrule
\multirow{7}{*}{\makecell[c]{Computer \\ Science}} & Kimi & 5.30 & 4.95 & 8.68 & 18.93 \\
& GLM & 8.05 & 6.32 & 8.76 & 23.13 \\
& Doubao & 8.05 & 6.67 & 8.90 & 23.62 \\
& Chengpian & 6.00 & 1.70 & 7.00 & 14.70 \\
& SurveyX & 7.78 & 6.06 & 8.53 & 22.37 \\
& SurveyGo & 8.26 & 6.83 & 8.60 & 23.69 \\
& ScienceOne & 8.30 & 7.28 & 8.97 & 24.55 \\
\midrule
\multirow{7}{*}{\makecell[c]{Six STEM \\ Disciplines}} & Kimi & 7.03 & 5.75 & 8.44 & 21.22 \\
& GLM & 5.83 & 1.78 & 6.72 & 14.33 \\
& Doubao & 5.42 & 5.03 & 8.92 & 19.37 \\
& Chengpian & 7.67 & 5.83 & 8.42 & 21.92 \\
& SurveyX & 7.64 & 5.86 & 8.58 & 22.08 \\
& SurveyGo & 7.81 & 5.86 & 8.50 & 22.17 \\
& ScienceOne & 7.78 & 6.78 & 9.00 & 23.56 \\
\bottomrule
\end{tabular}
\vspace{-1em}
\end{table}

\subsection{Experimental Setup}

We use the original survey title as the query for each automatic survey system to obtain its generated survey. All systems are capable of performing web retrieval, ensuring fair access to external information sources. We evaluate three categories of systems: 
general-purpose long-text writing systems 
(e.g., Kimi~\cite{Kimi_2025}, GLM~\cite{GLM_2025}), 
paper-writing systems 
(e.g., Chengpian~\cite{Chengpian_CP_2025}, Doubao Paper Mode~\cite{Doubao_2025}), 
and dedicated survey-generation systems 
(e.g., SurveyX~\cite{surveyx}, SurveyGo~\cite{surveygo}, ScienceOne~\cite{ScienceOne}). 
All systems receive the same input query, and their outputs are evaluated under identical criteria. 
The aggregated scores are reported below.

\subsection{Main Results}

\paragraph{Overall Evaluation.}
Table~\ref{tab:main_results} reveals distinct patterns across the three categories of systems. General long-text models show good language fluency and topic relevance, but their surveys remain shallow: originality, synthesis, and critical analysis are consistently weak. Paper-writing systems provide stronger structural control, yet their outputs are often rigid and template-driven, limiting analytical depth and reducing the diversity of perspectives. Dedicated survey-generation systems achieve the highest scores overall, with clearer taxonomies, more coherent synthesis, and noticeably better fact–opinion separation. Their advantage is particularly evident in STEM subjects, where effective retrieval and domain-aware structuring matter most. These differences indicate that high-quality survey writing requires more than general LLM capabilities—it benefits from specialized pipelines that explicitly target literature grounding and structured synthesis.

\paragraph{Outline Evaluation.}
Table~\ref{tab:outline_results} shows that dedicated survey-generation systems produce the most coherent outlines, with clearer hierarchical structure and more informative section titles. General long-text models generate fluent headings but struggle to maintain consistent, field-appropriate organization. Paper-writing systems perform unevenly: while structurally disciplined, some outputs become overly rigid or lack meaningful content signals. Overall, the results indicate that effective outline design—capturing both structure and scholarly intent—is a core strength of specialized survey systems and a common weakness of generic models.

\paragraph{Reference Evaluation.}
Table~\ref{tab:reference_results} shows a clear gap between the two systems capable of producing full reference lists. ScienceOne achieves both higher recall and higher precision, indicating stronger retrieval grounding and fewer hallucinated or irrelevant citations. SurveyX, despite covering a reasonable portion of the literature, introduces more incorrect or unmatched references. This contrast highlights that citation reliability—especially avoiding unsupported references—remains a key weakness for many systems.

\begin{table}[t]
\centering
\caption{Reference evaluation results.}
\label{tab:reference_results}
\begin{tabular}{lrrr}
\toprule
\multicolumn{1}{l}{\textbf{System}} & \multicolumn{1}{c}{\textbf{Recall}} & \multicolumn{1}{c}{\textbf{Precision}} & \multicolumn{1}{c}{\textbf{F1}} \\
\midrule
SurveyX & 76.85 & 75.09 & 75.96 \\
ScienceOne & 90.58 & 84.28 & 87.32 \\
\bottomrule
\end{tabular}
\end{table}

\section{Conclusion}
We introduced \textbf{SurveyEval}, a comprehensive benchmark for assessing LLM-generated academic surveys across content quality, outline coherence, and reference accuracy. SurveyEval provides a practical tool for analyzing and identifying limitations in current survey-writing systems, supporting more targeted improvements in future development.

% \begin{acks}
% \end{acks}

\bibliographystyle{ACM-Reference-Format}
\bibliography{sample-base}

@misc{surveygo,
      title={LLM$\times$MapReduce-V2: Entropy-Driven Convolutional Test-Time Scaling for Generating Long-Form Articles from Extremely Long Resources}, 
      author={Haoyu Wang and Yujia Fu and Zhu Zhang and Shuo Wang and Zirui Ren and Xiaorong Wang and Zhili Li and Chaoqun He and Bo An and Zhiyuan Liu and Maosong Sun},
      year={2025},
      eprint={2504.05732},
      archivePrefix={arXiv},
}

@misc{surveyx,
      title={SurveyX: Academic Survey Automation via Large Language Models}, 
      author={Xun Liang and Jiawei Yang and Yezhaohui Wang and Chen Tang and Zifan Zheng and Shichao Song and Zehao Lin and Yebin Yang and Simin Niu and Hanyu Wang and Bo Tang and Feiyu Xiong and Keming Mao and Zhiyu li},
      year={2025},
      eprint={2502.14776},
      archivePrefix={arXiv},
}

@misc{ScienceOne,
  author       = {Institute of Automation, Chinese Academy of Sciences},
  title        = {ScienceOne},
  year         = {2025},
  howpublished = {Online},
  url          = {https://scienceone.ia.ac.cn}
}

@misc{Chengpian_CP_2025,
  author       = {Baidu, Inc.},
  title        = {Chengpian},
  year         = {2025},
  howpublished = {Online},
  url          = {https://cp.baidu.com/}
}

@misc{Doubao_2025,
  author       = {ByteDance Ltd.},
  title        = {Doubao},
  year         = {2025},
  howpublished = {Online},
  url          = {https://www.volcengine.com/}
}

@misc{Kimi_2025,
  author       = {Moonshot AI},
  title        = {Kimi},
  year         = {2025},
  howpublished = {Online},
  url          = {https://kimi.ai/}
}

@misc{GLM_2025,
  author       = {Zhipu AI},
  title        = {GLM},
  year         = {2025},
  howpublished = {Online},
  url          = {https://open.bigmodel.cn/dev/api}
}

@misc{agarwal2024litllms,
  title        = {LitLLMs, LLMs for Literature Review: Are we there yet?},
  author       = {S. Agarwal},
  year         = {2024},
  eprint       = {2412.15249},
  archivePrefix= {arXiv},
  primaryClass = {cs.CL},
}

@misc{ali2024automatedLiteratureReview,
  title        = {Automated Literature Review Using NLP Techniques and Retrieval-Augmented Generation (RAG) with Large Language Models},
  author       = {N. F. Ali},
  year         = {2024},
  eprint       = {2411.18583},
  archivePrefix= {arXiv},
  primaryClass = {cs.CL},
}

@article{yan2025surveyforge,
  title={Surveyforge: On the outline heuristics, memory-driven generation, and multi-dimensional evaluation for automated survey writing},
  author={Yan, Xiangchao and Feng, Shiyang and Yuan, Jiakang and Xia, Renqiu and Wang, Bin and Zhang, Bo and Bai, Lei},
  journal={arXiv preprint arXiv:2503.04629},
  year={2025}
}

@inproceedings{2024autosurvey,
title={AutoSurvey: Large Language Models Can Automatically Write Surveys},
author = {Wang, Yidong and Guo, Qi and Yao, Wenjin and Zhang, Hongbo and Zhang, Xin and Wu, Zhen and Zhang, Meishan and Dai, Xinyu and Zhang, Min and Wen, Qingsong and Ye, Wei and Zhang, Shikun and Zhang, Yue},
booktitle={The Thirty-eighth Annual Conference on Neural Information Processing Systems},
year={2024}
}


\end{document}